\documentclass[table,xcdraw,x11names]{Interspeech}
\usepackage{tcolorbox}

\interspeechcameraready

\title{The State Of TTS: A Case Study with Human Fooling Rates}

\author[affiliation={1}]{Praveen}{Srinivasa Varadhan}
\author[affiliation={1}]{Sherry}{Thomas}
\author[affiliation={1}]{Sai}{Teja M S}
\author[affiliation={2}]{Suvrat}{Bhooshan}
\author[affiliation={1}]{Mitesh}{M. Khapra}

\affiliation{AI4Bharat}{Indian Institute of Technology Madras}{India}
\affiliation{}{Gan.AI}{India}
\email{cs21d201@cse.iitm.ac.in, miteshk@dsai.iitm.ac.in}
\keywords{speech synthesis, human-centric evaluation}

\usepackage{comment}
\usepackage{booktabs}
\usepackage[normalem]{ulem}  

\definecolor{customOrangePastel}{HTML}{f9dcc0}
\definecolor{customOrange}{HTML}{FFE0B2}
\definecolor{customPinkPastel}{HTML}{ffe5e7}
\definecolor{customPink}{HTML}{FCE4EC}
\definecolor{customBlue}{HTML}{E3F2FD}
\definecolor{deepBlue}{HTML}{0D47A1}
\definecolor{deepRed}{HTML}{C62828}
\definecolor{deepOrange}{HTML}{D84315}

\newcommand{\strikeadd}[2]{\textcolor{deepBlue}{\textbf{\small{\sout{#1} #2}}}}

\begin{document}

\maketitle

\begin{abstract}
While subjective evaluations in recent years indicate rapid progress in TTS, can current TTS systems truly pass a human deception test in a Turing-like evaluation?
    We introduce Human Fooling Rate (HFR), a metric that directly measures how often machine-generated speech is mistaken for human. Our large-scale evaluation of open-source and commercial TTS models reveals critical insights: (i) CMOS-based claims of human parity often fail under deception testing, (ii) TTS progress should be benchmarked on datasets where human speech achieves high HFRs, as evaluating against monotonous or less expressive reference samples sets a low bar, (iii) Commercial models approach human deception in zero-shot settings, while open-source systems still struggle with natural conversational speech; (iv) Fine-tuning on high-quality data improves realism but does not fully bridge the gap. Our findings underscore the need for more realistic, human-centric evaluations alongside existing subjective tests.
\end{abstract}

\section{Introduction} 

The gold standard for artificial intelligence has always been indistinguishability from humans, as exemplified by the Turing Test. In speech synthesis \cite{tan2021survey}, this means a TTS system must produce speech that is not just preferred in subjective evaluations but is truly indistinguishable from a natural speaker. If a system fools human listeners into believing they are hearing real speech, it has met the highest standard of evaluation. As AI-driven dialogue systems become more integrated into daily interactions, the demand for genuinely human-like synthetic speech has never been greater. With chatbots and virtual assistants becoming more lifelike, it is time to set a higher bar for evaluating speech synthesis—not just incremental MOS or CMOS improvements, but actual perceptual indistinguishability. 


Subjective evaluation tests such as CMOS \cite{Loizou2011}, MUSHRA \cite{mushra2015}, and MOS \cite{kirkland23_ssw,wester2015are} have long been effective in guiding TTS model development and should continue to do so. These metrics provide valuable insights into preference and quality, helping researchers iterate on models. However, as TTS systems are increasingly deployed in real-world applications, there is a need for an additional evaluation that directly measures whether synthetic speech is truly indistinguishable from human speech. Such a deployment-centric evaluation should be more direct and interpretable to overcome the limitations of existing tests \cite{wester2015are, chiang2023report,varadhan2024rethinking}, which often lack clear real-world implications. For example, consider a MUSHRA test where a system scores 86 and the reference 90, both labeled ``Excellent.'' Does this mean the system is ready for deployment? If a CMOS score surpasses that of a human reference, does that definitively indicate the system passes a human deception test? Our findings suggest this is not necessary.

To complement existing evaluation methods, we introduce Human Fooling Rate (HFR), a deployment-centric metric that directly measures how often machine-generated speech is mistaken for human. Unlike traditional subjective evaluation methods, HFR is not about preference or relative quality but deception: Can the listener confidently distinguish real from synthetic speech? 
We conduct a large-scale HFR evaluation of ten state-of-the-art TTS systems — 5 top-performing open-source models and 5 commercial offerings — engaging 135 participants across different experimental setups and voice conditions.  This evaluation is crowdsourced via Prolific, ensuring a diverse and representative listener pool for assessing perceptual indistinguishability at scale.


Our findings reveal several critical gaps in current TTS evaluation methods. \textbf{(Finding 1)} State-of-the-art TTS systems can achieve high CMOS/MUSHRA scores by closely matching the reference, yet still perform poorly on HFR tests. This suggests a reference-matching bias, where raters prioritize similarity over genuine naturalness. Additionally, subtle synthetic cues, such as, digital voice quality and artifacts, may be overlooked in preference-based evaluations but become evident in deception-based assessments. \textbf{(Finding 2)} A major concern is that many TTS evaluations use benchmarks where even reference human recordings have low HFR scores, as they sound monotonic and lack expressive variation. This allows TTS models to appear successful by matching suboptimal references rather than achieving true human-like speech, creating a false sense of progress. Meaningful evaluation requires benchmarks where human speech itself achieves high fooling rates, ensuring synthetic speech is judged against realistic perceptual standards. 
\textbf{(Finding 3)} Our results further show that while commercial models approach human deception in zero-shot settings, \textbf{(Finding 4)} open-source TTS systems continue to struggle with natural conversational speech, and fine-tuning on high-quality conversational data leads to only partial improvements. These insights underscore the need for more comprehensive evaluation frameworks, and we propose HFR as a crucial complement to existing metrics, offering a more robust and interpretable standard for TTS benchmarking.

\section{The Human Fooling Rate Test}

In this section, we introduce a complementary metric that evaluates perceptual indistinguishability rather than subjective preference, addressing limitations in existing evaluation methods.

\noindent \textbf{Definition.} \quad The Human Fooling Rate (HFR) is defined as the percentage of times machine-generated speech is mistaken for human speech in a binary forced-choice listening test. Mathematically, it is computed as:

\begin{equation}
\text{HFR} =  \sum_{i=1}^{N} \sum_{j=1}^{T} \frac {\mathbb{I} ( y_{i,j} = \text{human} )} {N \times T} \times 100
\end{equation}
where \( N \) is the total number of listeners, \( T \) is the total number of trials, \( y_{i,j} \) is the response of listener \( i \) for trial \( j \), and \( \mathbb{I} ( y_{i,j} = \text{human} ) \) is an indicator function that returns 1 if the listener labels the TTS speech as human and 0 otherwise.

\noindent \textbf{Procedure.} \quad In the HFR test, listeners are presented with individual speech recordings and must determine whether the audio is produced by a human speaker or a TTS system. To ensure fair evaluation, all participants are instructed to use headphones in a quiet environment and listen to each recording completely, without interruption,  before making a decision. While making their judgment, listeners are instructed to focus on key perceptual cues such as voice quality (e.g., robotic or compressed sound), unnatural modulation, monotonic delivery, inappropriate emotion or intonation, mispronunciations, skipped or repeated words, unnatural pauses or speed, and digital artifacts. By guiding listeners to consider these factors before making their decision, the evaluation process aims to ensure a more informed and reliable measure of a system’s ability to deceive human perception.


\section{Evaluation of State-of-the-Art TTS}
To assess whether state-of-the-art TTS systems can truly deceive human listeners, we systematically select models that represent the current landscape of speech synthesis. We describe our model selection criteria, benchmarks, evaluation design, and the evaluation platform used for conducting large-scale perceptual tests via crowd-sourcing.

\noindent\textbf{Model Selection.} 
In real-world applications, such as voice assistants, dubbing or accessibility tools, personalized and dynamic voice synthesis is increasingly essential, where users expect high-quality, speaker-adaptive TTS without requiring extensive training data. To meet this demand, we focus on speech prompt-based TTS models capable of zero-shot voice cloning, as they best align with real-world needs by enabling natural speech synthesis from minimal speaker input. Additionally, these systems can function as traditional TTS models by using training voices as prompts, ensuring strong performance on familiar speakers while retaining the flexibility for new voice adaptation. We evaluate both open-source and commercial TTS systems, selecting models that claim human-level synthesis (via CMOS or MUSHRA scores), release pretrained checkpoints, support fine-tuning, and perform well in existing TTS leaderboards. 
Based on these criteria, we select the following open-source models: StyleTTS2 \cite{Li2023styletts2}, XTTS \cite{casanova2024xtts}, GPT-SoVITS \cite{gptsovits2024}, F5-TTS \cite{Chen2024F5-TTS}, and VoiceCraft \cite{Pen2024Voicecraft}. For commercial TTS, we evaluate ElevenLabs \cite{elevenlabs_tts_2025} and PlayHT \cite{playht_tts_2025}. These models represent a strong baseline for assessing the deception capability of modern prompt-based TTS systems.

\noindent 
\textbf{Evaluation Design \& Benchmarks.} Our goal is to benchmark prompt-based TTS systems capable of voice cloning, where prompt quality plays a crucial role in output realism. \textbf{[Evaluation 1]} We begin by evaluating systems on three widely used benchmarks, viz., LJSpeech \cite{ljspeech17}, LibriTTS \cite{Zen2019LibriTTS}, and LibriSpeech \cite{Panayotov2015LibriSpeech}, to establish baseline deception rates. We synthesize outputs by randomly sampling speaker prompts from the respective test sets, ensuring that the prompt and target utterance are always distinct. Our findings prompt us to question the deception quality of these popular benchmarks. 

Building on the insights gained and hypothesizing that higher-quality voices could enhance deception rates, 
 we explore whether high-quality open-source voices can serve as effective prompts. Given Expresso's \cite{Nguyen2023Expresso} challenging nature with its natural and expressive conversational speech, we use it to determine (i) \textbf{[Evaluation 2]} if open-source voices can improve deception rates and (ii) \textbf{[Evaluation 3]} whether adaptation through fine-tuning can further enhance HFR scores. This systematic process enables a comprehensive assessment of model performance across diverse datasets, recording conditions, and evaluation settings, covering both zero-shot and fine-tuned scenarios.


\noindent \textbf{Evaluation Platform.}  We conduct evaluations on \textsc{SAFFRON}\footnote{\url{https://github.com/AI4Bharat/saffron}} (Speech Assessment Framework For Robust Objective and Normative Evaluation), a platform we designed for scalable perceptual evaluation of TTS systems. SAFFRON supports both HFR and MUSHRA tests, ensuring a standardized and reproducible framework for benchmarking speech realism. \textsc{SAFFRON} enforces strict listening conditions by requiring participants to hear full samples before responding, tracking response times to prevent rushed judgments, and integrating seamlessly with Prolific for large-scale crowd-sourced evaluations. With its scalable design and robust experimental controls, SAFFRON provides a reliable platform for speech synthesis research. We publicly release it to facilitate more rigorous and interpretable evaluations of TTS systems.

\noindent \textbf{Crowd-sourcing Participants.} We recruit 135 native US-English speakers from Prolific \cite{prolific_2024}, ensuring balanced age and gender demographics. Participants must be born and residing in the US, have English as their primary language, and be 18–60 years old and have a task acceptance rate of at least 99\% on Prolific. These constraints help ensure that evaluations reflect real-world native speaker perception. Across all tests, participants provide over 30,300 ratings, and for each experiment we ensure that every system receives ratings from at least 30 participants across 30 utterances.  All procedures were approved by the institute’s ethics review board and the total cost of conducting these experiments, including participant compensation, amounts to approximately £3,400.

\section{Key Findings}

We present our findings on the ability of state-of-the-art TTS systems to produce human-like speech based on large-scale HFR evaluations. The following sections explore whether open-source models achieve human deception in zero-shot settings, the reliability of existing TTS benchmarks, and how factors like the realism of the reference voices and fine-tuning impact performance.

\subsection{Has open-source TTS reached human-level quality?}
\label{sec:open_benchmarks}

\begin{table}[!h]
\setlength{\tabcolsep}{4pt}
\caption{Human Fooling Rates (HFR) of Open-Source TTS Systems on popular test sets. $^*$ indicates model has seen the benchmark during training. (95\% CI: min.=2.87; max.=3.27)}
\label{tab:hfr-tts-on-benchmarks}
\centering
\begin{tabular}{@{}lcccc@{}}
\toprule
\textbf{System} & \textbf{LJSpeech} & \textbf{LibriTTS} & \textbf{LibriSpeech} & $\mathbf{\mu}$ \\ \midrule
Human           & 78.33             & 73.33             & 70.67                & 74.11      \\
StyleTTS2       & $61.33^{*}$             & $45.67^{*}$             & 45.67                & 50.89      \\
F5-TTS          & 49.67             & 43.67             & 47.00                & 46.78      \\
XTTS            & 59.33             & 41.33             & 38.00                & 46.22      \\
GPT-SoVITS      & 41.00             & 31.33             & 41.67                & 38.00      \\
VoiceCraft      & 37.33             & 28.33             & 31.00                & 32.22      \\ \bottomrule
\end{tabular}
\end{table}

\noindent We first evaluate open source TTS systems in a zero-shot setup by prompting them with voices from 3 popular benchmarks as shown in Table \ref{tab:hfr-tts-on-benchmarks}. 
The human HFR scores in the first row of the table represent the deception rate of real human recordings. While one might expect it to be 100\%, in reality, even natural speech is occasionally misclassified as synthetic. This can be attributed to factors such as recording artifacts and variations in speaking style.
We see in Table \ref{tab:hfr-tts-on-benchmarks} that no open-source TTS system comes close to matching human recordings in fooling rates. Even the best-performing system, StyleTTS2 that claims a CMOS of +0.28 on LJSpeech only attains 61\% HFR on the same benchmark. Overall, it achieves only 50.89\% HFR against the human baseline (74.11\%). Likewise, F5-TTS reports a CMOS  of +0.31 on Seed-TTS test-en, yet scores a HFR of 46.78\% overall across benchmarks. This indicates that while synthesis quality has improved over time, truly indistinguishable speech remains an open challenge. 

\begin{tcolorbox}[colback=customOrange, colframe=customOrange, arc=2mm, boxrule=0mm, left=2pt, right=2pt, top=2pt, bottom=2pt]
\textbf{Finding \#1:} \textit{
Claims of near human parity based on CMOS can crumble under a Turing-like deception test, exposing the gap between perception and reality. This calls for using stronger complementary evaluation methods (like HFR) that directly test ``natural speech'' claims, ensuring assessments align with real-world human indistinguishability.
}
\end{tcolorbox}

Open-source TTS models claim strong generalization, yet their HFR scores vary significantly across datasets, indicating a lack of robustness. For example, XTTS achieves an HFR of 59.33\% on LJSpeech but drops to 38.00\% on LibriSpeech, suggesting poor zero-shot speaker generalization to more diverse test sets. LJSpeech consistently results in higher HFR scores, possibly because it presents an easier benchmark with less speaker variation or because it aligns more closely with the training data. In contrast, LibriSpeech yields the lowest HFR scores for human recordings, likely due to its diverse range of speakers and challenging recording conditions. However, is it not futile to expect that using prompts derived from such benchmarks can enable prompt-based TTS (whose goal is to mimic the input prompt accurately) to deceive humans better? In contrast, using prompts from such benchmarks for CMOS or MUSHRA tests sets an artificially low standard for TTS systems, as their simplistic style and limited variation make them easy to mimic. This can create a misleading impression of progress.

\begin{tcolorbox}[colback=customPink, colframe=customPink, arc=2mm, boxrule=0mm, left=2pt, right=2pt, top=2pt, bottom=2pt]
\textbf{Finding \#2: } \textit{
Benchmarks with challenging recording environments and more diverse speakers yield weaker deception in TTS, reinforcing the idea that systems trained or evaluated on such datasets may inherit their limitations rather than overcome them.}
\end{tcolorbox}


\subsection{How well do TTS systems fair on the high-quality Expresso Voices?}
Motivated by our earlier findings that low-quality prompts can lead to poor deception, we now examine another key aspect of prompt-based TTS: its ability to replicate high-quality voices in conversational settings rather than narration-style speech prevalent in the popular benchmarks covered in section \ref{sec:open_benchmarks}. We use Expresso which features professional voice actors delivering natural and expressive conversational speech, making it an ideal testbed. We prompt systems with two distinct voices (ex02 and ex03) and test their zero-shot ability to mimic humans. Additionally, we include two commercial systems, viz., ElevenLabs \cite{elevenlabs_tts_2025} and PlayHT \cite{playht_tts_2025} with instant voice cloning capabilities to determine whether closed-domain models can achieve high deception rates. 

{Table \ref{tab:hfr-expresso-zs} presents HFR scores on the Expresso benchmark, highlighting the clear distinction between closed-domain commercial systems and open-domain models.}
PlayHT (HFR: 71.49) and ElevenLabs (HFR: 69.85) achieve same deception rates as reference human audio samples (Human HFR: 70.68), whereas open-source models lag significantly behind. This suggests that state-of-the-art commercial models excel in zero-shot speaker adaptation to high-quality conversational speech, likely due to specialized training and access to proprietary high-fidelity datasets. Similar progress in both modeling and dataset quality may be required for open-source TTS to reach natural-sounding synthesis.

We also report MUSHRA scores in Table \ref{tab:hfr-expresso-zs}, which reaffirms that relative subjective metrics can inflate perceived realism \textbf{(Finding \#1)}. For example, XTTS scores a MUSHRA of 76.58 (higher than Human) yet only fools listeners 41.8\% of the time. This indicates that while listeners may rate audio quality highly in MUSHRA, they can still detect subtle cues (such as digital artifacts) that expose its synthetic origin in HFR.
Therefore, relying solely on CMOS-like or MUSHRA evaluations may overestimate naturalness, reinforcing the need for complementary deception-based metrics like HFR that measure a system’s ability to truly mimic human speech.

\begin{table}[!h]
\caption{HFR on Expresso (95\% CI: min.=4.04; max.=4.45)}
\label{tab:hfr-expresso-zs}
\centering
\begin{tabular}{@{}llcc@{}}
\toprule
\textbf{System} & \textbf{Domain} & \textbf{HFR} & \textbf{MUSHRA} \\ \midrule
PlayHT          & \textit{Closed} & 71.49        & \cellcolor[HTML]{F7FCFE}85.37           \\
Human           & \textit{Ref.}   & 70.68        & 74.78           \\
ElevenLabs      & \textit{Closed} & 69.85        & \cellcolor[HTML]{F7FCFE}80.39           \\
F5-TTS          & \textit{Open}   & 50.26        & 70.75           \\
GPT-SoVITS      & \textit{Open}   & 44.61        & 68.21           \\
XTTS            & \textit{Open}   & 41.80        & \cellcolor[HTML]{F7FCFE}76.58           \\
StyleTTS2       & \textit{Open}   & 38.60        & 71.21           \\
VoiceCraft      & \textit{Open}   & 30.52        & 49.02           \\ \bottomrule
\end{tabular}
\end{table}

\begin{tcolorbox}[colback=customBlue, colframe=customBlue, arc=2mm, boxrule=0mm, left=2pt, right=2pt, top=2pt, bottom=2pt]
\textbf{Finding \#3:} \textit{In the zero-shot setting, commercial models are able to achieve parity with human speech, whereas open-source TTS systems far lack in generating natural conversational speech.}
\end{tcolorbox}

\subsection{Does fine-tuning on high-quality voices improve deception?}
Given that open-source systems struggle to achieve high deception rates on unseen voices, it is interesting to assess their performance when trained on high-quality seen voices. To validate this, we fine-tune the best (F5-TTS) and worst (VoiceCraft) performing open-source models (in the zero-shot setting) and measure their HFR scores before and after training. Table \ref{tab:hfr-expresso-ms} shows that while fine-tuning boosts fooling rates, it does not fully close the gap. F5-TTS improves marginally while VoiceCraft sees a larger jump upto 43.45\%. This suggests that exposure to higher-quality data helps, but fine-tuning on this 40 hour dataset alone seems insufficient to reach human deception levels. Perhaps, more data or better training recipes and architecture are required in the open-source.

\begin{table}[!h]
\caption{Human Fooling Rates of systems after fine-tuning on the Expresso Benchmark. (95\% CI: min.=4.67; max.=4.51)}
\label{tab:hfr-expresso-ms}
\centering
\begin{tabular}{@{}lcc@{}}
\toprule
\textbf{System}     & \textbf{Zero-Shot} & \textbf{Many-Shot} \\ \midrule
\textbf{F5-TTS}     & 50.26              & \textbf{52.22}     \\
\textbf{VoiceCraft} & 30.52              & \textbf{43.45}     \\ \bottomrule
\end{tabular}
\end{table}

\begin{tcolorbox}[colback=customBlue, colframe=customBlue, arc=2mm, boxrule=0mm, left=2pt, right=2pt, top=2pt, bottom=2pt]
\textbf{Finding \#4:} \textit{Fine-tuning on high-quality conversational voices provides modest gains in fooling rates, and open-source TTS systems still fall short of the standard for human deception. Achieving truly natural speech in the open-source may require larger datasets, improved training strategies, or fundamental model enhancements.}
\end{tcolorbox}

\subsection{Why do open-source TTS Systems score high on MUSHRA but low on HFR?}

\begin{table}[!h]
\centering
\setlength{\tabcolsep}{4pt} 
\caption{\% of times each marker was identified by raters as indicative of machine-generated speech, comparing Human, Commercial, and Open-source systems on Expresso (ex02).}
\label{tab:hfr-g-expresso-ex02}
\begin{tabular}{@{}lccc@{}}
\toprule
\textbf{Marker}                    & \textbf{Human}                       & \textbf{Commercial}                  & \textbf{Open-source}         \\ \midrule
Voice Quality is Digital. & \cellcolor[HTML]{FFFEFE}\textbf{6.9} & \cellcolor[HTML]{FFFCFD}9.3          & \cellcolor[HTML]{FCE4EC}36.1 \\
Unnatural pauses.                  & \cellcolor[HTML]{F7FCFE}\textbf{4.0} & \cellcolor[HTML]{FFFEFF}6.7          & \cellcolor[HTML]{FEF0F5}22.8 \\
Unnatural pitch.                   & \cellcolor[HTML]{FFFFFF}5.8          & \cellcolor[HTML]{FFFFFF}\textbf{5.6} & \cellcolor[HTML]{FEF5F8}17.2 \\
Flat or monotonic.                 & \cellcolor[HTML]{EDF9FE}\textbf{2.2} & \cellcolor[HTML]{FEFEFE}5.1          & \cellcolor[HTML]{FEF2F6}20.6 \\
Inappropriate emotion.             & \cellcolor[HTML]{F2FAFE}\textbf{3.1} & \cellcolor[HTML]{F3FBFE}3.3          & \cellcolor[HTML]{FFFAFC}11.4 \\
No human quirks.                   & \cellcolor[HTML]{F0FAFE}2.7          & \cellcolor[HTML]{EDF9FE}\textbf{2.2} & \cellcolor[HTML]{FFFAFC}11.4 \\
Mispronunciations.                 & \cellcolor[HTML]{E1F5FE}\textbf{0.2} & \cellcolor[HTML]{E1F5FE}\textbf{0.2} & \cellcolor[HTML]{FFFBFD}9.8  \\
Word skips/repeats.                & \cellcolor[HTML]{E3F5FE}0.7          & \cellcolor[HTML]{E2F5FE}\textbf{0.4} & \cellcolor[HTML]{FFFDFE}7.6  \\
Digital artifacts.                 & \cellcolor[HTML]{E2F5FE}\textbf{0.4} & \cellcolor[HTML]{E3F5FE}0.7          & \cellcolor[HTML]{FFFFFF}5.2  \\ \bottomrule
\end{tabular}
\end{table}

\noindent To better understand why open-source TTS systems achieve high MUSHRA scores yet low HFR values, we conducted a granular HFR test on the Expresso benchmark (ex02) with 15 raters. In this test, participants were not only asked to determine whether speech was machine-generated but also to specify the particular flaws that led them to that conclusion. These flaws were labelled by selecting one or more specific error markers from a predefined list of nine, as enlisted in Table \ref{tab:hfr-g-expresso-ex02}. The results in Table \ref{tab:hfr-g-expresso-ex02} reveal that the most common giveaway for open-source models being identified as machine is their digital voice quality (36.1\%), followed by unnatural pauses (22.8\%) and flat or monotonic delivery (20.6\%). In contrast, commercial models exhibit error rates similar to or better than human recordings in key areas like pitch variation, human-like quirks (e.g., natural breaths), and reduced word skips or repeats. These findings explain the stark gap in deception performance between open-source and commercial systems in Table \ref{tab:hfr-expresso-zs}, despite their comparable MUSHRA scores. More importantly, this granular HFR analysis provides targeted insights for improving TTS models beyond overall quality metrics, highlighting key areas for advancing open-source synthesis toward true perceptual indistinguishability. Although HFR is designed as a deployment-centric metric, its granular version offers detailed diagnostic feedback, making it a valuable development-centric tool too for model refinement.

\subsection{Are HFR tests more efficient?}

In Table \ref{tab:hfr-mushra-time-taken}, we see that the average time taken per audio sample is significantly lower for HFR tests compared to traditional MUSHRA evaluations. Notably, the granular HFR test, which provides targeted insights into specific artifacts, is completed less than half the time of MUSHRA. This remarkable efficiency, combined with rich diagnostic feedback, makes HFR tests a powerful complement to conventional perceptual evaluations. 

\begin{table}[!h]
\centering
\setlength{\tabcolsep}{4pt} 
\caption{Average duration (s) taken per listener to rate one audio sample per system. 
}
\label{tab:hfr-mushra-time-taken}
\begin{tabular}{@{}lccc@{}}
\toprule
\textbf{}           & \textbf{HFR}                                         & \textbf{HFR-Granular}                                & \textbf{MUSHRA}                                      \\ \midrule
\textbf{Time Taken} & \cellcolor[HTML]{FFFDF9}{\color[HTML]{000000} 24.30} & \cellcolor[HTML]{FFFFFF}{\color[HTML]{000000} 22.53} & \cellcolor[HTML]{FFE0B2}{\color[HTML]{000000} 42.45} \\ \bottomrule
\end{tabular}
\end{table}

\section{Related Work}

Subjective relative assessments such as MOS, CMOS, and MUSHRA have been widely used in TTS evaluation but have faced substantial criticism \cite{wester2015are, kirkland23_ssw, varadhan2024rethinking, lemaguer2024limits}. MOS tests are known to be highly variable \cite{finkelstein2023importance}, context-sensitive \cite{clark2019evaluating}, and prone to biases like range-equalization \cite{cooper2023investigating}. MUSHRA assessments too are susceptible to reference-matching bias and judgment ambiguity \cite{varadhan2024rethinking}. Several works, slightly modify tests \cite{shen2024naturalspeech, Lajszczak2024basetts} or propose new ones \cite{kayyar2023subjective} to potentially overcome limitations. Given these limitations, deception-based evaluations inspired by the Turing Test \cite{turing1950computing} offer a compelling alternative. While such evaluations have shown promise in NLP \cite{uchendu2021turingbench, jones2024people}, they remain under-explored in TTS. Our work bridges this gap by introducing HFR, a direct deception-based measure for evaluating machine-generated speech.

\section{Limitations}
Like all subjective evaluations, HFR is not immune to variability in ratings, test design biases, or perceptual differences among raters. While it shifts the focus from preference-based comparisons to a deception-based evaluation, it still inherently relies on human perception. Owing to budget constraints, we limit our experiments to a subset of top open-source and commercial models, leaving room for broader validation across other TTS systems. We emphasize that HFR is not a replacement for CMOS/MUSHRA but a complementary metric that provides a deployment-centric perspective on TTS evaluation.

\section{Conclusion}
As TTS systems continue to advance, the ultimate test of progress should not be limited to preference-based evaluations but must also address perceptual indistinguishability from human speech. Our large-scale HFR evaluation reveals that even top-performing systems struggle to fully deceive human listeners.  
Existing TTS benchmarks often overestimate system performance by failing to reflect real-world human deception rates.
While commercial models show promise in zero-shot settings, open-source TTS lags behind, with fine-tuning on high-quality data offering only limited gains. These findings emphasize the need for stronger evaluation frameworks that go beyond traditional MOS and CMOS scores.  By introducing HFR, we aim to provide a deployment-centric metric that directly measures human-likeness, paving the way for more rigorous benchmarking and future advancements in speech synthesis.

\section{Acknowledgements}
We thank Digital India Bhashini, MeitY, Government of India, for generously supporting this work. We are grateful to the EkStep Foundation and Nilekani Philanthropies for their grant enabling critical human and cloud resources. We also thank Google for supporting Praveen through the Google Ph.D. Fellowship. Finally, we thank the Prolific platform for facilitating our crowd-sourced evaluations.



\bibliographystyle{IEEEtran}
\bibliography{refs, refs_rethinking}

\end{document}